\pdfoutput=1

\documentclass[11pt]{article}

\usepackage[final]{acl}

\usepackage{times}
\usepackage{latexsym}

\usepackage[T1]{fontenc}

\usepackage[utf8]{inputenc}

\usepackage{microtype}

\usepackage{inconsolata}

\usepackage{graphicx}

\usepackage{printlen}

\usepackage{float}
\usepackage{booktabs}
\usepackage[export]{adjustbox}
\usepackage{pgfplots}
\usepackage{tikz}
\usetikzlibrary{matrix}
\pgfkeys{/pgf/number format/.cd,fixed,precision=3}
\pgfplotsset{
    compat=1.3, 
    every axis/.append style={scale only axis, axis on top,
    height=4.5cm, width=5cm, xmin=-1, xmax=240,
    }
}
\usetikzlibrary{patterns}
\usepgfplotslibrary{statistics}

\DeclareRobustCommand{\legenddotA}{%
  \tikz[baseline=-0.5ex] {\draw[blue,pattern=dots,color=blue] (0,0) circle (1ex);}
}
\DeclareRobustCommand{\legenddotB}{%
  \tikz[baseline=-0.5ex] {\draw[red,pattern=dots,color=green] (0,0) circle (1ex);}
}
\DeclareRobustCommand{\legenddotC}{%
  \tikz[baseline=-0.5ex] \draw[red,pattern=dots,color=red] (0,0) circle (1ex);
}
\DeclareRobustCommand{\legenddotD}{%
  \tikz[baseline=-0.5ex] \draw[green,pattern=horizontal lines, color=violet] (0,0) circle (1ex);
}

\newcommand{\mysubsection}[1]{\vspace{0.3em}\noindent\textbf{#1}}

\title{Examining the Utility of Self-disclosure Types for Modeling\\Annotators of Social Norms}

\author{Kieran Henderson$^\spadesuit$ \and Kian Omoomi$^\heartsuit$ \and \\ \textbf{Vasudha Varadarajan$^\clubsuit$ \and Allison Lahnala$^\heartsuit$ \and Charles Welch$^\heartsuit$} \\
    $^\spadesuit$ University of Toronto, $^\clubsuit$ Carnegie Mellon University, $^\heartsuit$ McMaster University \\
    \texttt{hende325@cs.toronto.edu,vvaradar@andrew.cmu.edu,\{omoomik,lahnalaa,cwelch\}@mcmaster.ca}
}

\begin{document}
\maketitle
\begin{abstract}
Recent work has explored the use of personal information in the form of persona sentences or self-disclosures to improve modeling of individual characteristics and prediction of annotator labels for subjective tasks. The volume of personal information has historically been restricted and thus little exploration has gone into understanding what kind of information is most informative for predicting annotator labels. In this work, we categorize self-disclosures and use them to build annotator models for predicting judgments of social norms. We perform several ablations and analyses to examine the impact of the type of information on our ability to predict annotation patterns. 
Contrary to previous work, only a small number of comments related to the original post are needed. Lastly, a more diverse sample of annotator self-disclosures did not lead to the best performance. Sampling from a larger pool of comments without filtering still yields the best performance, suggesting that there is still much to uncover in terms of what information about an annotator is most useful for verdict prediction.
\end{abstract}

\section{Introduction}

Standard modeling practices rely on aggregated annotations, often suppressing disagreement and variation as noise.  
However, annotation disagreements are often not merely random errors, but rather systematic signals of differing perspectives rooted in individual characteristics \cite{basile2021we, cabitza2023toward}. 
This has given rise to perspectivist NLP, a growing movement that aims to preserve and model subjective diversity instead of resolving it through majority vote or adjudication \cite{frenda2024perspectivist}.

Subjectivity is central to many NLP tasks, especially \textit{social} NLP tasks—including hate speech detection, moral acceptability, social norms, irony detection, and stance classification—yet such tasks are often treated as though there were a single objective truth. These tasks are inherently subjective because they require interpretation through the lens of individual beliefs, values, and experiences~\cite{rottger-etal-2022-two}. Furthermore, language itself is a vehicle for expressing these diverse perspectives. The choice of words, tone, and even syntactic structures can encode subtle stances and attitudes, making the annotation and modeling of such tasks particularly challenging and subjective~\cite{,vijjini-etal-2024-socialgaze}. Aggregating these judgments into a single label risks erasing minority viewpoints and oversimplifying the rich diversity of human interpretation.

\begin{figure}
    \centering
    \includegraphics[width=\linewidth]{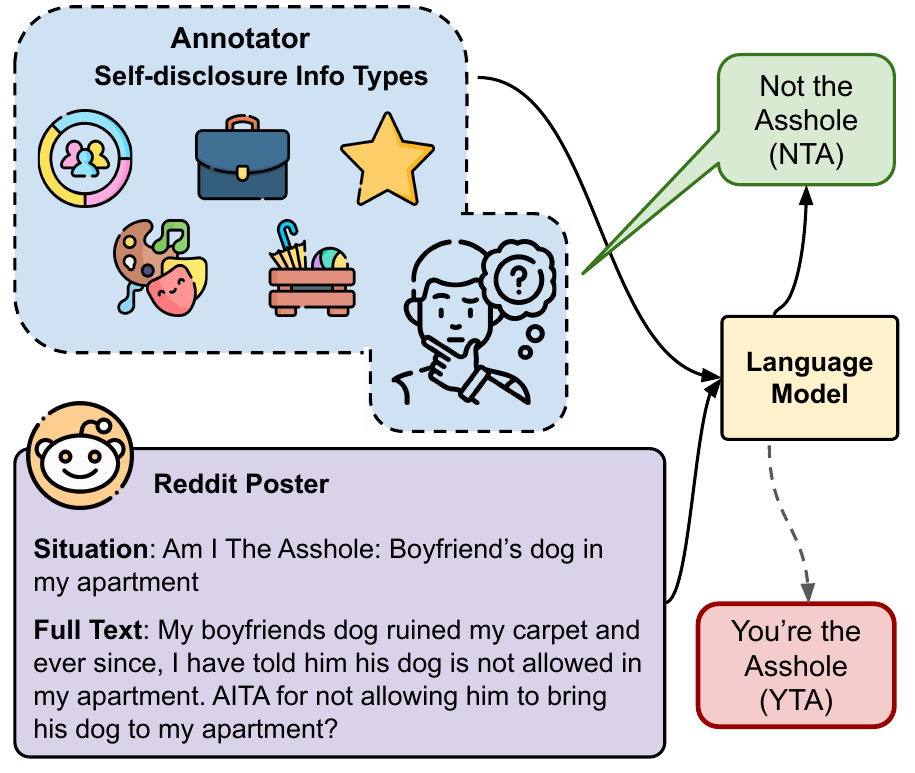}
    \caption{Illustration of the experimental setup testing how types of self-disclosure information about annotators (e.g., demographics, experiences, attitudes) influence a language model’s ability to predict their judgments in social dilemmas. The model uses Reddit posts as training data and an embedding of commenter self-disclosures to make a decision.}
    \label{fig:enter-label}
\end{figure}
In this work, we propose modeling subjective judgments of individual annotators by enriching the background context to include personal attributes derived from self-disclosure statements. These open-ended statements, which describe individuals' hobbies, possessions, professions, relationships, and attitudes, allow us to go beyond demographic categories to represent people more holistically. We conduct a systematic analysis of the types of self-disclosures on the subreddit \texttt{/r/amitheasshole}, using theoretical and automatic methods of categorizing self-disclosures, to test what kind of personal information is most useful for helping to predict the judgment of a commenter.

Our work addresses three research questions. \textbf{RQ1.} What strategies and parameters for sampling information about annotators produces the most effective models of an annotators judgments of social situations? Specifically, we examine 1) how many samples are ideal for learning a person's judgment style, 2) whether samples that are semantically similar to the social situations offer more predictive value than randomly selected ones, and 3) whether it is more effective to use selected sentences rather than entire comments. 
\textbf{RQ2.} How do different types of self-disclosures compare in how they influence the performance of annotator models?  We used theory based regular expressions to categorize self-disclosure types into demographics (e.g., age, gender), experiences (e.g., hobbies, possessions, work life), attitudes (e.g., opinions), and relationships (e.g., family members, relationship status). 
\textbf{RQ3.} How do automatic categorizations of self-disclosure types compare to theory-based categorizations? We derive categorizations of self-disclosure types by data-driven clustering methods, and compare to our theory-based taxonomy.

Overall, our work addresses the gap in our understanding of the factors that influence an individual’s annotation behavior that remain underexplored \cite{fleisig-etal-2024-perspectivist,orlikowski-etal-2023-ecological}. By broadening the perspectivist agenda to encompass richer identity representations, we enhance ongoing efforts to model human subjectivity more accurately and to create systems that are personalized, equitable, and sensitive to human variability.

\begin{table*}[]
    \small
    \centering
    \begin{tabular}{lp{125mm}}
        \toprule
        \textbf{Disclosure Type} & \textbf{Examples} \\
        \midrule
        \includegraphics[width=15pt,valign=c]{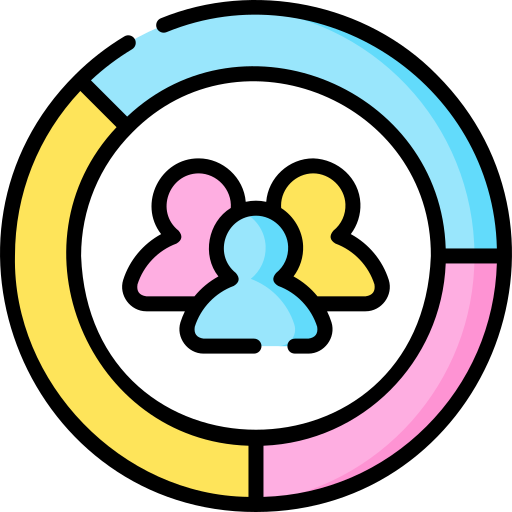} Demographics & \includegraphics[width=15pt,valign=c]{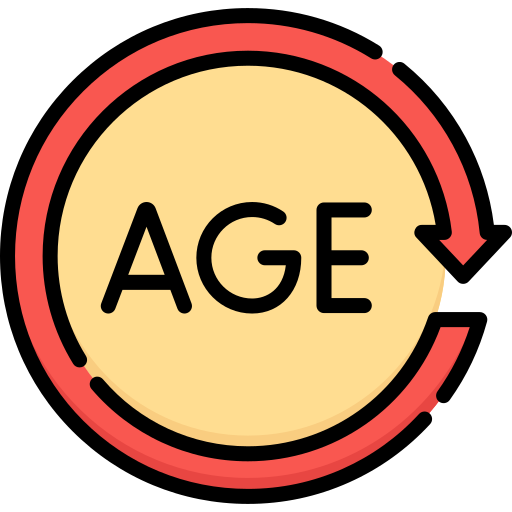} Age: I'm 22 yrs old and my mom is telling everyone that she isn't spending alot of money on \newline \phantom{12345678} Christmas this year. \newline \includegraphics[width=15pt,valign=c]{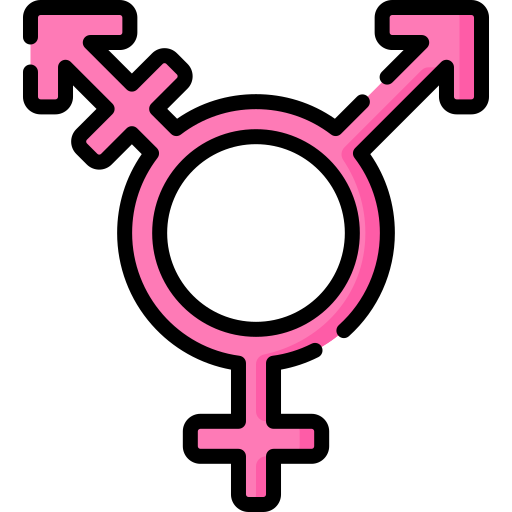} Gender: I'm an 100\% cis woman totally comfortable in my gender identity \newline \includegraphics[width=15pt,valign=c]{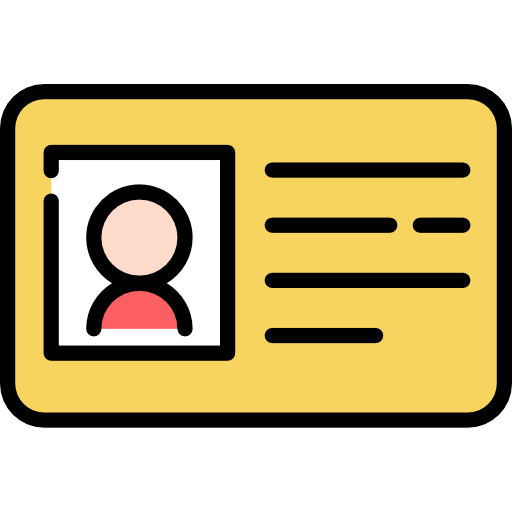} Identity: I'm an omnivore but I make food for myself that happens to be vegan \newline \phantom{12345678901} I'm happily married and attractive thanks\\
        \includegraphics[width=15pt,valign=c]{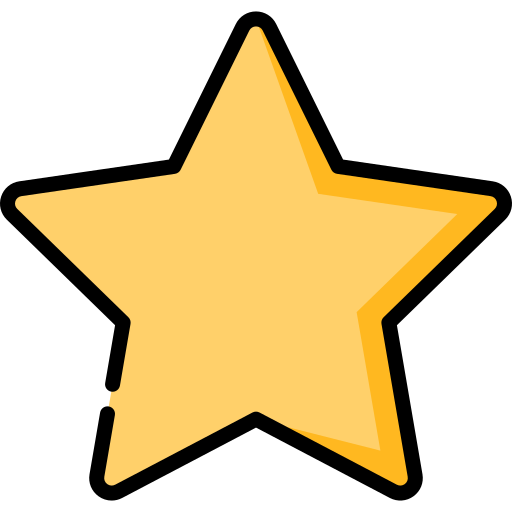} Experiences & \includegraphics[width=15pt,valign=c]{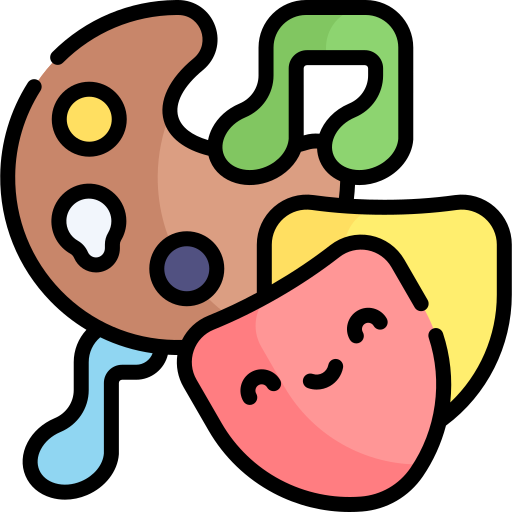} Hobby: I like to play video\textbackslash board games and have no friends and am super awkward \newline \includegraphics[width=15pt,valign=c]{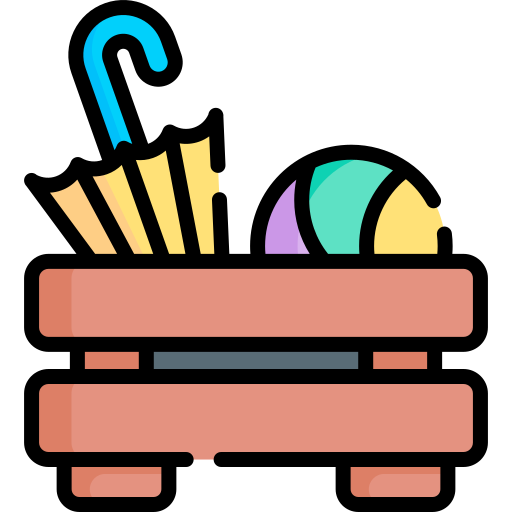} Possessions: I have five cats and they love to watch the cat and nature shows on youtube \newline \includegraphics[width=15pt,valign=c]{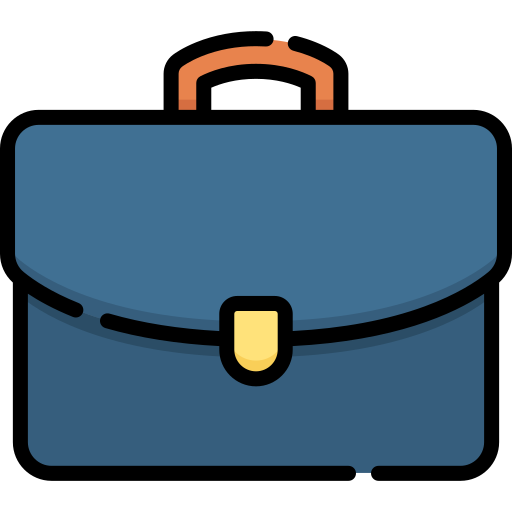} Work: I work as a civil engineer and the salary is decent but definitely not enough to be shelling\newline \phantom{123456789} out \$60K for a master's program \\
        \includegraphics[width=15pt,valign=c]{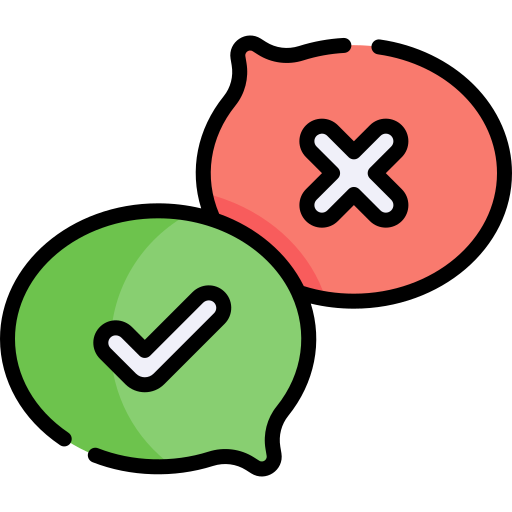} Attitudes & I think it’s ridiculous that people aren’t allowed to use computers in tests in this day and age \newline I consider American policing one of the most authoritarian parts of my government \\
        \includegraphics[width=15pt,valign=c]{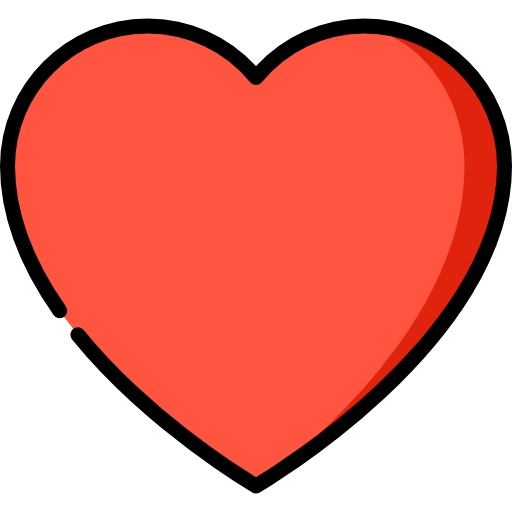} Relationships & I got the feeling that my sister \& friends think everything I have was handed to me or came easily. \newline I have a friend I have known for a couple of years now, she lives in another country but we have seen each others a couple of times and we talk daily and I would say she is a very good friend of mine. \\
        \bottomrule
    \end{tabular}
    \caption{List of self-disclosure types, displaying the high-level types (left column), and lower-level types for demographics and experiences (right-column) along with example sentences in our corpus.}
    \label{tab:disclosure_types}
\end{table*}

\section{Background}\label{sec:related_work}

\paragraph{Perspectivism}Many NLP pipelines rely on annotated corpora, where the quality of the annotations is measured by the agreement between multiple annotators. Traditionally, 
disagreements may be considered noise, signs of ill-defined tasks, or poor annotator training; therefore, a single ground truth is typically derived by various aggregation techniques. However, disagreements are common and often systematic, and for many problems there is no single ground truth~\cite{basile2021we}. Recent perspectivist NLP approaches reconsider the value of information and knowledge that can be derived from diverging annotations \cite{cabitza2023toward}, especially for highly complex or subjective tasks.

Particularly in subjective tasks, multiple valid perspectives may exist, making the use of a single ground truth problematic. 
When considering highly subjective tasks like offensive or toxic language detection, individual differences in demographics and personality significantly influence how people label dimensions of the data \cite{leonardelli-etal-2021-agreeing,sang2022origin,sap-etal-2022-annotators,pei-jurgens-2023-annotator,frenda-etal-2023-epic}. Perspectivist approaches to these tasks not only help mitigate issues like silencing minority opinions in model representations, but also improve the predictive precision and robustness of models \cite{cabitza2023toward}, as was shown in research on hate speech detection \cite{akhtar2020modeling}, and irony detection \cite{frenda-etal-2023-epic}. 
Even in “seemingly objective tasks” \cite{basile2021we,frenda2024perspectivist}, like coreference resolution \cite{recasens2011identity} or part-of-speech tagging \cite{plank-etal-2014-linguistically}, nuanced models with a continuum/variety of perspectives may more effectively capture the linguistic phenomena.
Perspectivism in NLP, therefore, is a growing movement encouraging the construction and distribution of disaggregated datasets to embrace multiple, valid perspectives, and evaluation against individual labels rather than aggregated gold standards~\cite{frenda2024perspectivist}.

One approach to perspectivism is to learn from the disagreements themselves. \citet{fornaciari-etal-2021-beyond} present one way to do this, by predicting a soft-label distribution, or the distribution of annotator responses, as an auxiliary task for part-of-speech tagging and stemming, finding that it leads to improved performance.
One approach is to model individuals as annotators directly rather than collapsing their judgments into a single aggregated label. For example, \citet{golazizian-etal-2024-cost, lee-goldwasser-2022-towards} introduced the Moral Foundations Subjective Corpus and proposed a multitask architecture that captures general task-level patterns and preserves differences among annotators. Their approach uses a few-shot sample selection technique to capture how new annotators’ labeling behaviors differ from the general patterns, which enables the learning of individual perspectives. Similarly, \citet{davani-etal-2022-dealing} developed multi-annotator models with varying degrees of shared parameters, finding that explicitly accounting for annotator specific signals can match or exceeded aggregation based methods on subjective tasks such as hate speech detection and emotion classification, while also producing more informative uncertainty estimates.

Another approach is to group annotators by shared characteristics and study perspective variance with respect to grouped variables. An increasing number of works demonstrate that a significant portion of disagreement in subjective annotation tasks can be explained by demographic factors. For example, \citet{pei-jurgens-2023-annotator} identified systematic variance associated with annotators' age, gender, race, and education levels in tasks such as question-answering, offensiveness, and politeness rating. Similarly, \citet{sap-etal-2022-annotators} operationalize annotator attitudes related to political and linguistic ideologies, showing that variation in language toxicity judgments is rooted in these ideological measures in addition to demographic factors such as gender and race. 
However, \citet{fleisig-etal-2024-perspectivist} point out that demographics are often used to explain some amount of annotator disagreement, but other factors may better account for differences in perspectives. Along similar lines, \citet{deng-etal-2023-annotate} analyzed eight datasets and found that only a limited portion of annotator variance can be explained by observable attributes such as age, gender, geographic location, household income, political leaning, and education. 
\citet{orlikowski-etal-2023-ecological} explicitly incorporated sociodemographic factors into multi-annotator models by extending the architecture of \citet{davani-etal-2022-dealing} with group level layers. Their results showed no significant performance gains, suggesting that perspective variation may be driven more by individual differences than by coarse demographic groupings.

\citet{plepi-etal-2024-perspective} looked at using the most similar persona sentences to a given post (to which a verdict will be assigned) generate responses that consider individual perspectives. They extended the data from \citet{forbes-etal-2020-social} to collect a larger set of comments from the AITA subreddit that allows them to experiment more with selecting the best self-disclosure statements. They find that choosing the top 15-20 most similar sentences worked the best for their use case.

Recent work on automated detection of self-disclosures has leveraged taxonomies based on the sensitivity or ambiguity of the personal information revealed. For example,
\citet{bak-etal-2014-self} propose a self-disclosure topic model, categorizing conversational turns in social media as high (sensitive information), medium (non-sensitive information), or no self-disclosure.
\citet{barak2007degree} define levels according to the degree of personal thoughts, feelings, and information in a statement. 
\citet{valizadeh-etal-2021-identifying} built a classifier using levels for ambiguity or uncertainty of disclosure of symptoms in a medical setting. 

Unlike previous works, 
our framework centers on the \textit{content} of disclosures that is immediately relevant to the annotator. 
Specifically, we introduce a taxonomy grounded in Social Penetration Theory, 
and focus on disclosures in online forum posts rather than conversational turns.
This allows us to analyze the disclosures themselves using content features understood to be salient to annotators, 
and to examine how the depth and nature of personal information disclosed can enhance our characterization of annotator experience and perspective.

 \paragraph{Social Penetration Theory}
This sociopsychological theory
conceptualizes the development of interpersonal relationships as a function of self-disclosure: the degree of intimacy between individuals increases as the information shared goes from superficial to increasingly personal details, in both a one-to-one and group setting~\cite{altman1973reciprocity, carpenter2015social}. 
According to this framework, individuals reveal themselves in layers, beginning with superficial information such as \textbf{Demographics} (age, occupation, identity or background), followed by more personal disclosures, including \textbf{Attitudes} (e.g., political or social views), \textbf{Past Experiences }(such as significant life events or formative memories), and details about \textbf{Relationships} (such as family dynamics or close friendships). 
These categories of self-disclosures provide a structured lens for analyzing how a person's self-concept, beliefs, and life history may influence their judgments and interactions, offering a nuanced framework for modeling annotator behavior in interpersonal or evaluative contexts. 
In this paper, we aim to expand our understanding of the types of self-disclosures that systematically influence judgments in subjective tasks, including not only sociodemographic identities but also information about lived experiences from self-disclosure statements about hobbies, professions, belongings, attitudes, and interpersonal relationships.

\section{Data}

We use a set of posts from the Reddit community \texttt{/r/amitheasshole} (AITA). In the AITA forum, Reddit users post a description of a social situation and the actions they took and ask if they have done something wrong. Community members provide their judgments in the form of YTA (you're the asshole) or NTA (not the asshole). 
We use the data from \citet{plepi-etal-2024-perspective} which was derived from \citet{welch-etal-2022-understanding}'s dataset of AITA situations (which in turn used a set of posts crawled by \citet{forbes-etal-2020-social} for the \texttt{Social-Chem-101} dataset) by filtering for authors that have between 20 and 500 additional comments somewhere else on the subreddit. \citet{plepi-etal-2024-perspective}'s resulting dataset contains 21k posts written by 15k authors, and 212k verdicts provided by 13k annotators. These annotators also have an additional 3.6M comments from other parts of the subreddit excluding the 21k posts. The distribution is skewed toward NTA, with around 70\% of all verdicts being NTA for all splits. This may be because people prefer not to use the YTA label, or possibly because posters are biased in their portrayal of themselves.
Each post has a title (situation description), full post text, author, and a set of comments from other Reddit users (commenters/annotators) which each provide a verdict (judgment YTA/NTA) and a justification or additional comment to supplement the verdict. 

Our paper performs an annotator modeling task, wherein the aim is to predict how a user would ``annotate'' their verdict label given what we know about the author from self-disclosure statements. 
Collecting deeper self-disclosures as described by social penetration theory~\cite{tang2012self}, is often expensive and challenging. While AITA posts are not formally ``annotated'' but rather ``participated in'' with some incentive, the platform’s anonymity and reduced real-world consequences make individuals more willing to share personal information~\cite{de2014mental, miller2020investigating}. This makes AITA a practical proxy for large-scale annotation, offering a more affordable and relatively high-quality alternative.

\begin{table}[]
    \small
    \centering
    \begin{tabular}{lccc}
        \toprule
        \textbf{Cluster} & \textbf{No. Posts} & \textbf{No. Authors} \\
        \midrule
        C1: Financial Issues & 42,773 & 7,772\\ 
        C2: Acc. Social Needs & 91,111 & 9,870\\ 
        C3: Parenting \& Discipline & 50,183 & 8,211\\ 
        C4: Family Struct. Change & 50,182 & 8,058\\ 
        C5: Rel. Issues w/Men & 64,779 & 9,591\\ 
        C6: Rel. Issues w/Women & 83,321 & 10,378\\ 
        C7: Sympathy \& Support & 35,912 & 8,436\\ 
        C8: Family Disputes & 50,512 & 8,600\\ 
        C9: Negative Affect & 75,743 & 9,311\\ 
        C10: Food \& Meals & 25,003 & 6,113\\ 
        \bottomrule
    \end{tabular}
    \caption{The number of posts and unique authors in each of the clusters}
    \label{tab:cluster_post_counts}
\end{table}

To investigate how different types of self-disclosures influence annotation modeling, we first sought to categorize annotator comments for the types of self-disclosures they contain.
We attempted to assign both theory-based and automatic category assignments to each comment in our dataset to represent the type of information disclosed by annotators. 
Our theory-based assignment was informed in part by our intuition and related psychological theory, and in part by our exploratory analysis of the automatic assignments. 

\subsection{Automatic Categorization}
We started with both LDA topics~\cite{blei2003latent} and k-means clusters as ways of grouping comments. When clustering and topic modeling, we use only posts that contain a self-disclosure from the list in Appendix~\ref{sec:appendix_sd_sentences} so that clusters relate more to personal statements. This leaves us with 570k posts, or 16\% of the original data. About 400 authors have never written a post with a self-disclosure phrase and were excluded from our category experiments. Clusters were created using post embeddings generated using SBERT with the same parameters discussed in \S\ref{sec:methods}. Both approaches provide a method by which a single group label (topic or cluster) can be assigned to each comment. We found through manual inspection in preliminary experiments that both provided reasonable groupings and decided to proceed with k-means, as it is more easily interpretable. We also experimented with changing the number of clusters by multiples of five up to 25, finding that ten provided a reasonably interpretable and coherent set of clusters.

\begin{table}[]
    \small
    \centering
    \begin{tabular}{lccc}
        \toprule
        \textbf{Theory Based Category} & \textbf{No. Posts} & \textbf{No. Authors} \\
        \midrule
        Attitudes & 254,755 & 12,415\\ 
        Demographics & 211,292 & 11,211\\ 
        Experiences & 153,064 & 10,195\\ 
        Relationships & 132,964 & 10,640\\ 
        \bottomrule
    \end{tabular}
    \caption{The number of posts and unique authors in each of the four high level theory based groups}
    \label{tab:regex_post_counts}
\end{table}

We determined a label for each cluster by examining a random sample of posts from each cluster as well as a small set of posts close to the cluster centroid. The clusters were labeled by two of the authors separately and subsequently adjudicated. This resulted in the following clusters: (1) financial problems, (2) accommodating social needs, (3) parenting, often relating to discipline, (4) changing family structure, cutting off ties to some, marrying or having children, (5) relationship issues with men, (6) relationship issues with women, (7) sympathy and support for other's behavior, (8) family disputes, that largely appear intergenerational, (9) negative affect, or taking issue with the someone's actions/opinion, or behavior that is unacceptable, and lastly, (10) meal-related conflicts.

\begin{table*}[]
    \centering
    \small
    \begin{tabular}{lcccccccccccc}
        \toprule
        \textbf{Number of Samples} & \multicolumn{2}{c}{5} & \multicolumn{2}{c}{10} & \multicolumn{2}{c}{15} & \multicolumn{2}{c}{20} & \multicolumn{2}{c}{25} & \multicolumn{2}{c}{30} \\
        & Acc & F1 & Acc & F1 & Acc & F1 & Acc & F1 & Acc & F1 & Acc & F1\\
        \midrule 
        Random Comments        & 67.1 & 58.1 & 67.7 & 58.7 & 68.6 & 57.5 & 68.6 & 59.1 & 67.3 & 58.4 & 68.3 & 58.7 \\
        Random Sentences        & 67.1 & 57.8 & 66.3 & 57.5 & 68.0 & 58.9 & 67.4 & 58.1 & 67.2 & 58.1 & 67.2 & 57.7 \\
        Similar Comments        & \textbf{71.4} & \textbf{62.6} & 69.7 & 60.1 & 70.5 & 60.9 & 69.2 & 60.0 & 67.9 & 58.5 & 68.8 & 59.2 \\
        Similar Sentences       & 67.4 & 57.4 & 67.4 & 57.8 & 67.8 & 58.2 & 67.7 & 57.4 & 68.5 & 57.8 & 68.6 & 57.0 \\
        \bottomrule
    \end{tabular}
    \caption{Situation split performance (Accuracy and F1) across varying numbers of maximum samples, with best performance in bold. Similar 5 comments to random 5 comments statistically significant (permutation test, $p<0.0002$).}
    \label{tab:samples——sit}
\end{table*}

\subsection{Theory-based Categorization}\label{sec:theory-based-categorization}

Building upon the Social Penetration Theory for self disclosure~\cite{altman1973reciprocity}, 
we introduce a theory-based categorization framework for understanding the types of self-disclosures that occur during interpersonal communication.
Our taxonomy of annotator self-disclosure statements includes (a) \textbf{Demographics}, which was found to be the most common in other studies on annotator modeling, (b) \textbf{Experiences}, which includes generally self-descriptive statements about one's past experiences and relatively more stable traits, (c) \textbf{Attitudes}, which are self-descriptive statements of social views, morals, values and belief systems, and (d) \textbf{Relationships}, which describe the sociability and familial relations of the annotator. 
The clustering methods in all preliminary experiments picked up on the relationship between those in conflict, similar to what was found in \citet{welch-etal-2022-understanding}. 
We analyzed the most common n-grams around self-disclosure phrases, grouped authors by post count, tagged parts of speech to identify frequent patterns and verbs, and used Stanza’s coreference resolution \citep{stanza} to clarify sentence meaning and better understand the main themes in self-disclosure.
Theory-based and automatic approaches group self-disclosures by the type of information and the context (e.g. affect or conflict type) in which the disclosure was made, respectively.
Post and author counts for each category are listed in Tables~\ref{tab:cluster_post_counts} and \ref{tab:regex_post_counts}.
We extended the age and gender regular expressions from \citet{welch-etal-2020-compositional} and \citet{plepi-etal-2022-unifying} to include a variety of ways of expressing age, non-binary and transgender, along with age and gender combinations, e.g. (24F representing 24 year old female). Our set of initial self-disclosure statements was expanded from \citet{plepi-etal-2024-perspective} and similar to \citet{mazare-etal-2018-training}, but without enforcing part-of-speech constraints. We also borrowed expressions from \citet{tigunova-etal-2020-reddust}, which included age, gender, family status, profession, and hobby, but expanded the set of cases. The full set of regular expressions we developed is listed in Table~\ref{tab:regex} in the Appendix. 

After using regex to extract the self-disclosure information, we randomly selected 100 comments from each group to annotate by hand. We found that there were 9 comments in the demographic group, 17 posts in the attitudes group, 25 in the experiences group, and 11 in the relationships group which were not clean self-disclosure phrases. For example, the attitudes regex patterns currently capture the phrases: ``I think driving is scary'' and ``I think thats a good plan''. The first phrase gives a clear opinion and is fully contained. The second sentence gives an opinion on something that has been mentioned outside of the current comment. We were initially expecting to extract clear opinions with this pattern, but in reality self-contained disclosures are difficult to extract. Note that the use of an LLM could potentially help, but would introduce another source of error, while requiring substantially more compute effort for a large corpus. Future work could likely improve upon our findings by improving the extraction approach for self-disclosures, possibly through a multi-stage filtering process.

We list the full set of disclosure types in Table~\ref{tab:disclosure_types}. There are four high-level categories, with demographics and experiences each having lower-level categories. Demographics includes age, gender, and other identity statements, while experiences covers hobbies, possessions, and occupations. Attitudes includes opinions that match something like ``I think,'' ``I feel,'' or ``I believe.'' Relationships captures familial and friend relationships.
After filtering the original dataset, we are left with a total of 1.1M posts, representing 31\% of the original set of annotator comments. We provide an analysis of unigram and bigram statistics in Appendix~\ref{sec:self_disclosure_exploration}.

\section{Methodology}\label{sec:methods}

As a baseline, we reproduced the results of \citet{plepi-etal-2022-unifying} for their situation split using the same parameters from their setup. They showed that the situation split was the most challenging, i.e. predicting how annotators respond to new conflict posts (see Appendix~\ref{sec:appendix_other_splits} for details on other splits). We used DistilRoBERTa \cite{sanh2019distilbert} for the base SBERT classifier~\cite{reimers-gurevych-2019-sentence} with a hidden size 768 and max length 512. We use an added linear layer for classification, trained the model for 10 epochs with the Adam optimizer, learning rate $1e-3$, and focal loss~\cite{lee-etal-2021-towards}.

\mysubsection{Experimental Setup.} The brief exploration of self-disclosure statements in previous work on generation raised several questions for us. \textbf{RQ1:} Which sampling parameters perform best? First, how many samples are ideal for classifying judgments? Second, are similar samples more useful than random samples? Lastly, is selecting sentences more effective than taking the whole posts? We hypothesized that comments similar to the post we are classifying should perform better. Given the results of \citet{plepi-etal-2024-perspective}, we assumed that 15-20 similar posts would work best. However, we hypothesized that individual sentences may work better because they focus on only the important information and eliminate potential noise in longer posts.

Next, we explored how the different types of personal information categories affect performance. \textbf{RQ2:} How do the theory-based self-disclosure types compare in their impact on performance? Due to the sparsity of data for the lower-level categories for demographics and experiences (see Table~\ref{tab:disclosure_types}), we performed this experiment with only the four high-level categories. We hypothesized that identities (demographics) would outperform other categories because they most directly described who the person was. We put attitudes next, as this encompasses a persons views. We ranked experiences third and relationships last, thinking that someone's history and what they have learned would more greatly impact judgments than who they were related to.

Our next question was whether or not manually categorizing personal information was necessary, or if we could rely on automatic methods. \textbf{RQ3:} Does automatically categorizing self-disclosures via clustering perform as well as theory-based categorization? We hypothesized that clusters would underperform theory-based categories because clustering makes it difficult to enforce that similarity among clustered comments is driven mainly by author related information. rather than by topical or surface level features. Furthermore, we hypothesized that the best representation of an individual should include a diverse array of information about that person. If we sample from different manual and automatic categories, we expect performance to increase.

To perform the first set of experiments we used all comments from all annotators, meaning that some randomly sampled or similar comments/sentences may not belong to any theory-based or automatic category. We chose to investigate the sampling parameters first (RQ1), as the search space grows exponentially if we combine this with RQ2 and RQ3. For the latter research questions, we sampled only similar posts from one category at a time. For each category, we use the same set of annotators, however, each annotator does not have the same number of comments in each category, so we used up to five.

All of the models in our experiments were trained using a single NVIDIA H100 GPU. Generating similar comment embeddings and sentence tests took approximately 5 hours and only needed to be computed once each. Generating the embeddings for each category took less than a minute. A new model was needed for each of our experimental settings, a total of 24 models for the experiments in Table~\ref{tab:samples——sit}, a total of 14 models for the experiments in Table~\ref{tab:situtation_results}. Each model took approximately 2 hours to train and the results were averaged over 5 runs.

\begin{table}[]
    \small
    \centering
    \begin{tabular}{lccc}
        \toprule
        \textbf{Annotator Model} & \textbf{5+\%} & \textbf{Accuracy} & \textbf{F1} \\
        \midrule
        \multicolumn{4}{c}{Baseline Tests} \\ 
        \midrule
        No Comments & 0 & 67.0 & 56.0 \\
        All Comments & 100 & 65.9 & 55.9 \\
        \midrule
        \multicolumn{4}{c}{Theory Based High-level Categorization} \\ 
        \midrule
        Attitudes  & 68 & 66.1 & \textbf{57.4} \\ 
        Demographics  & 69 & 67.4 & \underline{58.6} \\ 
        Experiences & 48 & 66.7 & \textbf{56.1} \\ 
        Relationships & 56 & \underline{67.8} & \textbf{56.0} \\ 
        \midrule
        \multicolumn{4}{c}{Automatic High-level Categorization} \\ 
        \midrule
        C1: Financial Issues & 30 & 67.3 & 56.2 \\ 
        C2: Acc. Social Needs & 42 & 67.3 & 56.1 \\ 
        C3: Parenting \& Discipline & 33 & \underline{67.7} & \textbf{57.2} \\ 
        C4: Family Struct. Change & 33 & 66.9 & 56.6 \\ 
        C5: Rel. Issues w/Men & 37 & 64.8 & \textbf{55.8} \\ 
        C6: Rel. Issues w/Women & 42 & 66.8 & 55.8 \\ 
        C7: Sympathy \& Support & 24 & 66.5 & \textbf{\underline{57.9}} \\ 
        C8: Family Disputes & 32 & \underline{67.7} & \textbf{56.6} \\ 
        C9: Negative Affect & 39 & 66.8 & 57.3 \\ 
        C10: Food \& Meals & 23 & 65.6 & \textbf{56.3} \\ 
        \midrule
        \multicolumn{4}{c}{Diverse Sampling} \\ 
        \midrule
        Both Categorizations & - & \underline{68.1} & \textbf{56.0} \\ 
        Theory Based Categorization & - & 67.6 & \textbf{\underline{56.7}} \\ 
        Automatic Categorization & - & 66.3 & \textbf{55.6} \\ 
        \bottomrule
    \end{tabular}
    \caption{
    Accuracy and F1 scores for annotator models trained on the baseline, theory based, automatic, and diverse grouping methods. 5+\% indicates the percentage of annotators who had at least 5 posts for the category. All comments baseline represents previous work by \citet{plepi-etal-2022-unifying}. Bold numbers are statistically significant in comparison to the no comments baseline (permutation test, $p<0.05$), highest numbers per subsection are underlined.
    }
    \label{tab:situtation_results}
\end{table}

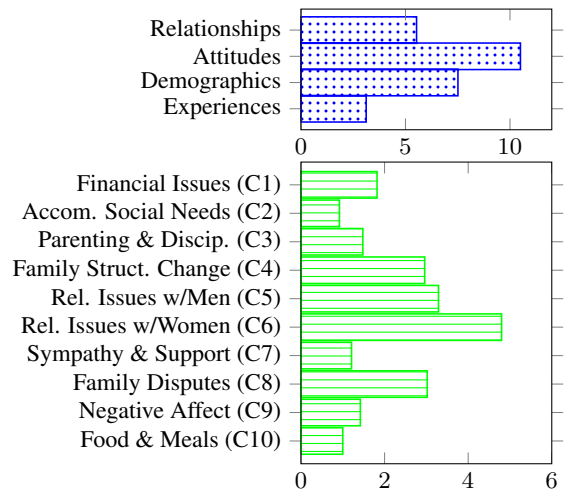
\begin{figure}[]
    \centering
    \small
    \begin{tikzpicture}
    \matrix{
    \begin{axis} [
        xbar,
        height=1.6cm,
        width=3.3cm,
        ymin=1, ymax=4,
        xmin=0, xmax=12,
        enlarge y limits = {abs = .8},
        ytick={1, 2, 3, 4, 5},
        yticklabels={Experiences, Demographics, Attitudes, Relationships},
    ]
    \addplot[draw = blue,
    	semithick,
    	pattern = dots,
    	pattern color = blue]
    coordinates {
    	(3.12,1) 
    	(7.51,2) 
    	(10.5,3)
    	(5.53,4) 
    };
    \end{axis}

    \\
    \begin{axis} [
        xbar,
        height=4cm,
        width=3.3cm,
        ymin=2, ymax=11,
        xmin=0, xmax=6,
        enlarge y limits = {abs = .8},
        ytick={2, 3, 4, 5, 6, 7, 8, 9, 10, 11},
        yticklabels={
                     Food \& Meals (C10),
                     Negative Affect (C9),
                     Family Disputes (C8),
                     Sympathy \& Support (C7),
                     Rel. Issues w/Women (C6),
                     Rel. Issues w/Men (C5),
                     Family Struct. Change (C4),
                     Parenting \& Discip. (C3),
                     Accom. Social Needs (C2),
                     Financial Issues (C1)},
    ]
    \addplot[draw = green,
    	semithick,
        pattern = horizontal lines,
    	pattern color = green
        ]
    coordinates {
        (1,2)
        (1.42,3)
        (3.02,4)
        (1.21,5)
        (4.8,6)
        (3.29,7)
        (2.96,8)
        (1.48,9)
        (0.92,10)
        (1.82,11)
    };
    \end{axis}
    \\
    };
    \end{tikzpicture}
    \caption{Percentage of times that each similar sentence falls into each theory based category (top, \legenddotA) and automatic cluster (bottom, \legenddotB) for experiments in Table~\ref{tab:samples——sit}. Most similar annotator comments do not belong to a category, as 75\% of similar posts have no theory based category and 78\% have no automatic category.}
    \label{fig:category_frequency}
\end{figure}

\section{Results}\label{sec:results}


The results of our experiments to determine the optimal strategies and parameters for sampling information about annotators (RQ1) are displayed in Table~\ref{tab:samples——sit}, showing accuracy and macro F1 score. We find that training on sentences underperforms using the full comments. It could be that after a certain point, adding more sentences or comments stops adding useful information. When the texts are sorted by similarity, there is a trade-off, where adding more similar text results in adding less relevant information. When increasing the number of similar sentences, we are adding a small amount of similar information, however the results from the model using 20 similar sentences already show a drop in performance because each sentence is less similar and less useful than the last. Overall, we find that adding the 5 most similar comments performs best. Using similar information, not surprisingly, is more effective than adding random posts or sentences. Using 5 similar comments significantly outperforms a baseline which uses all posts and which uses no posts (t-test, $p<0.0001$). We use the setting of 5 most similar comments for subsequent experiments in Table~\ref{tab:situtation_results}.

After identifying the optimal sampling parameters, we proceeded to address RQ2 and RQ3. Our results for experiments with different categories are shown in Table~\ref{tab:situtation_results}, with highest accuracy and F1 in bold separately for theory-based, automatic, and diverse categories. The highest performing theory-based category was demographics, followed by attitudes, and finally experiences, and relationships. We compared each result to a baseline that uses no comments from the annotators and a baseline that uses all comments, as in previous work. 
Unsurprisingly, no comments performs worst, while all comments achieves stronger performance while using substantially more data. 
The highest performing automatic clusters were C7: Sympathy \& Support and C9: Negative Affect. C3: Parenting \& Discipline had a notably high accuracy. Note that some of the higher performing categories were not significantly different. 
Notably, C10 had one of the lowest coverage rates with only 23\% of authors having five or more posts in that cluster. Diverse sampling performed competitively, with theory-based categorization outperforming automatic clusters or a combination of the two.

\pgfplotsset{width=7.5cm,height=5cm,compat=1.18}
\usepgfplotslibrary{statistics}
\begin{figure}[t]
\centering
\begin{tikzpicture}
\matrix{
\begin{axis}[boxplot/draw direction=y,
            name=plot1,
            ymin=0, ymax=100, 
            xmin=0.5, xmax=1.5,
            xtick={1}, xticklabels={},
            width=2.7cm, height=2.5cm]
    \addplot [boxplot prepared={
    lower whisker=0.269, lower quartile=19.512,
    median=33.566, upper quartile=50,
    upper whisker=94.737},
    ] coordinates {};

    \node at (axis cs: 1,28) {33.6};
\end{axis}

&
\begin{axis}[boxplot/draw direction=y,
            name=plot1,
            ymin=0.5, ymax=2, 
            xmin=0.5, xmax=1.5,
            xtick={1}, xticklabels={},
            width=2.7cm, height=2.5cm]
    \addplot [boxplot prepared={
    lower whisker=1, lower quartile=1,
    median=1.2, upper quartile=1.33,
    upper whisker=1.818},
    ] coordinates {};

    \node at (axis cs: 1,1.1) {1.2};
\end{axis} \\
};
\end{tikzpicture}
\caption{Distribution plots showing the percent coverage of author comments (left) in the top 5 most similar comments selected from all comments (Table~\ref{tab:samples——sit}) and the ratio between first and second rank counts of most frequently selected similar comments (right). Medians are printed below the median line.}
\label{fig:boxes_similar_posts}
\end{figure}
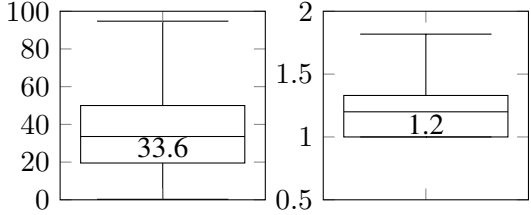

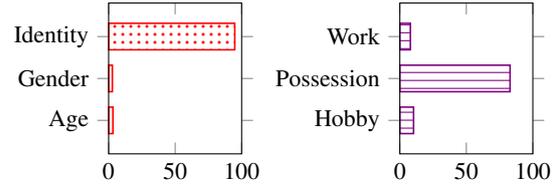
\begin{figure}
    \centering
    \small
    \begin{tikzpicture}
    \matrix{
    \begin{axis} [
        xbar,
        height=2cm,
        width=1.75cm,
        ymin=1, ymax=3,
        xmin=0, xmax=100,
        enlarge y limits = {abs = .8},
        ytick={1, 2, 3},
        yticklabels={Age, Gender, Identity},
        xtick={0, 50, 100},
        xticklabels={0, 50, 100},
    ]
    \addplot[draw = red,
    	semithick,
    	pattern = dots,
    	pattern color = red]
    coordinates {
    	(3.22,1) 
    	(2.78,2) 
    	(94.83,3) 
    };
    \end{axis}

    &
    \begin{axis} [
        xbar,
        height=2cm,
        width=1.75cm,
        ymin=1, ymax=3,
        xmin=0, xmax=100,
        enlarge y limits = {abs = .8},
        ytick={1, 2, 3},
        yticklabels={Hobby, Possession, Work},
        xtick={0, 50, 100},
        xticklabels={0, 50, 100},
    ]
    \addplot[draw = violet,
    	semithick,
        pattern = horizontal lines,
    	pattern color = violet
        ]
    coordinates {
        (10.32,1)
        (82.87,2)
        (7.92,3)
    };
    \end{axis}
    \\
    };
    \end{tikzpicture}
    \caption{Proportion of times in the training data that one of the top 5 most similar posts was selected from each subcategory, separately for demographics (left, \legenddotC) and experiences (right, \legenddotD), for top two rows of Table~\ref{tab:situtation_results}.}
    \label{fig:subcat_frequency}
\end{figure}

\section{Analysis and Discussion}\label{sec:discussion}

For \textbf{RQ1}, we hypothesized that a larger number of comments would work best and that sentences would be for more focused than comments. We found that similar samples significantly outperformed random samples.
In contrast to previous work, a smaller number of comments were needed. It appears that sentences, while possibly more focused, may be more highly variable, whereas comments provide a more substantial amount of information. It may also be the case that generation, as performed in \citet{plepi-etal-2024-perspective}, requires more examples to mimic style, whereas fewer samples are needed to predict verdicts, given that those samples are related. We examined the correlation between the length of comments in tokens and accuracy for 20 similar comments, finding a very weak Pearson $r=0.01$ but significant ($p<0.04$) correlation, suggesting the comment length has little to do with how accurately verdicts are classified. 

As mentioned in \S\ref{sec:results}, even a large number of sentences still underperforms a small number of similar posts. If we consider similar sentences to be only the unit of information that contains a fact about the annotator, it is more similar to work on personalization that uses personal attributes, as discussed in \S\ref{sec:related_work}. This type of metadata contains an age or gender, whereas the longer posts from annotators may contain rich latent information that perspectivist classifiers can leverage.
While it is not strictly the case that sentences from our corpus represent isolated facts, this reasoning supports, and may partially explain why comments are more useful than sentences. It also helps address the issue that sentences by themselves, are not always self-contained, as mentioned in \S\ref{sec:theory-based-categorization}.

Next, we found that our ranking of theory-based categories was correct (\textbf{RQ2}). Demographics was the highest performing category followed by attitudes, experiences, and lastly relationships. 
Lastly, for \textbf{RQ3}, we found that some automatic categories performed as well as theory-based categories, but that theory-based approaches (mostly demographics and attitudes) outperformed most clusters. It was not as intuitive to understand why one cluster outperforms another, e.g. why parenting and discipline (C3) outperforms relationship issues (C5 \& C6).
Similar amounts of effort went into constructing both theory-based and automatic categories. Future work will need to run more extensive experiments to find more comprehensive answers to these questions, but it appears that defining self-disclosure types using both approaches can lead to improved performance.

\mysubsection{Sampling Analysis.} We further examine the type of similar posts used to represent the annotators. For our experiments using all annotator comments, we plot the proportion of times that the five most similar comments used to represent that annotator come from each category, shown in Figure~\ref{fig:category_frequency}. We note that 75\% of these top-5 similar comments have no theory-based category and 78\% of them have no automatic category. Using either of our methods to categorize these posts eliminates the majority of similar comments. This explains the slightly lower performance when using comments from only one category rather than sampling from the full pool of posts from that author (as in Table~\ref{tab:samples——sit}.

We also looked at the proportion of times that posts were sampled from the low-level theory-based categories. These are shown in Figure~\ref{fig:subcat_frequency}. We see that similar posts most often select identity statements, rather than gender and age when using demographics. When we look at experiences, we see that most similar posts come from the set of possessions, with a lower proportion coming from work and hobbies. These proportions roughly follow the frequency of these types of posts.

We further examined the five most similar comments to assess the diversity of the sampled comments and the degree of imbalance in the distribution of the most frequently retrieved comments. 
Figure~\ref{fig:boxes_similar_posts} shows the relative frequency of the most frequent similar comment to the second most frequent similar comment with a median ratio of 1.2. This suggests that a diverse set of comments are used to represent the annotators across different situations.
We also see a median of 33\% of an authors comments end up in the set of similar comments at some point.

We further observed that the theory-based category, demographics, was the highest performing category. This group contains identities, which use the most generic regular expressions to match self-disclosures, making the group more diverse. 
Experiments from Table~\ref{tab:samples——sit} show that diverse sampling with theory-based categories is often outperformed by individual categories, both theory-based and automatic. As shown in Figure~\ref{fig:category_frequency}, we see that a variety of comments are sampled in these diverse experiments.
Diverse sampling significantly outperforms the \textit{all posts} method from previous work, but not the method that uses similar comments from the full set.
This evidence suggests that a diverse sample of information from an individual may improve performance, but also that we may still not be capturing the most useful information about individuals, as the similar comment set is much larger than what we are able to extract with either categorization approach.

\section{Conclusions}

We explored experimental settings for predicting verdicts based on self-disclosure statements. We found that contrary to prior work, only a small number of similar comments are needed to effectively model annotators and that in contrast, using many or all posts from an annotator performed worse. When examining the types of information that are most beneficial, we found that theory-based categories performed similarly to automatic categories (clusters), but that theory-based categories were more consistently high performing. 
Demographics outperformed all other types of self-disclosures. While our categories showed notable differences in performance, they all underperformed the five similar comments sampled from the full pool of annotator comments, suggesting that we are still not capturing the most useful information predicting verdicts.
The Reddit corpus allows us to explore the use of information that is not normally collected in the annotation process. We find that modeling certain kinds of information can be more informative when modeling individual annotators, suggesting that collecting a wider array of information about annotators during the annotation process could lead to ability to develop improved models.
We release our code and data to support future work on annotator modeling.\footnote{\url{https://github.com/KieranHenderson/Self-disclosuresForAnnotatorModeling}}

\section*{Limitations}

Self-disclosure statements were extracted with regular expressions that will not catch all cases of statements of a given type. They will also capture false positives. There are ways to improve the rule-based extraction of self-disclosures, though we found ours had a low enough false-positive rate to be useful for our experiments. One example of a false positive is with the category of possessions, which captures the pattern ``I have X'' but X could be ``had concerns,'' or otherwise describe the past rather than a present object or experience. Rules with parsing features could make this more specific.

We did not account for the change in attributes over time. If a poster has posted over many years, their age or gender identity would be more likely to change leading to a set of conflicting statements. These types of issues could be resolved with rules, e.g. by taking the most recent values, however the most recent values may not correspond to their thoughts when providing a verdict. One could argue that a bag of identity statements is more representative of an individual (i.e. they were active posting at a range of ages, gender identities), however, we are unaware of the magnitude of the impact of rule-based filtering of these statements.

It is difficult to account for all factors in a fully controlled experiment due to the interdependent nature of this kind of real world data. The number of annotators with at least five posts belonging to all automatic and theory-based categories was only 537 or 2.69\% of the total author set. For this reason, we cannot use the same number and type of available self-disclosures for each condition. When an annotator has less than five comments in a given category, we use the full available set and the similarity function has effectively no impact, as there are no posts to selectively exclude. These criteria may help future researchers in designing new data collection protocols.

Lastly, we have only examined the impact of these self-disclosures on Reddit and only in the judgment of social norms within one subreddit community. It remains to be seen how these findings generalize outside of this domain. The main barrier to testing this is the lack of available corpora with both annotations and sufficient self-disclosures.


\section*{Ethical Considerations}

Any work that attempts to model individual's, their opinions, language patterns, or behaviors has the potential to be misused. Modeling individual views could be used adversarially to find ways to present conflict situations that do not incite opposition, or conversely, how to present conflict that maximizes opposition for a target individual or group. 

There is the possibility that someone responsible for surveying or making decisions based on the views of a population would incorrectly assume they could use this type of modeling in place of directly interacting with said population, taking human beings out of the loop. Our data comes from Reddit, and while we attempted to include a diverse population and diverse collection of self-disclosures, the demographics of Reddit skew toward younger male and western individuals, and other demographics, ages, and genders, are underrepresented \cite{Studying_Reddit,pew_reddit}. We discourage the use of models built on this data for substituting any process that would normally involve humans in providing feedback.

\section*{Acknowledgments}

The icons used in this paper's figures and tables were provided by Freepik at flaticon.com.

\bibliography{custom}

\appendix

\section{Self-disclosure Statements}
\label{sec:appendix_sd_sentences}

Table~\ref{tab:regex} shows all regular expressions we used for the eight theory based categories of self-disclosure statements, all four high-level categories, plus three for each of demographics and experiences.

\begin{table*}[]
    \centering
    \begin{tabular}{rp{13cm}}
    \toprule
    \textbf{Category} & \textbf{Regex} \\
    \midrule
    Identity & \begin{minipage}{13cm}\begin{verbatim}\b(I am|I'm|Im)\s*(a|an)?\s*(.+?)(?=[,.!?]|$)\end{verbatim}\end{minipage} \\
    \midrule
    Gender & \begin{minipage}{13cm}\begin{verbatim}\b(?:I am|I'm|Im|I'?m)\s*(?:a|an)?\s*
(?P<sentence_gender>(?:non[-\s]?binary|male|female|man|woman|
boy|girl|guy|dude|mother|father|sister|brother|son|daughter|
husband|wife|trans(?: man| woman|gender|masculine|feminine)?|
genderfluid|agender|demiboy|demigirl|bigender|pangender))\b |
(?<!\w)(?P<age>\d{2})(?P<short_gender>(M|F|NB|AMAB|AFAB|MtF|FtM|
Enby|GM|GNC|TG|T|Cis))(?!\w)|
(?<!\w)(?P<short_gender2>(M|F|NB|AMAB|AFAB|MtF|FtM|Enby|GM|GNC
|TG|T|Cis))(?P<age2>\d{2})(?!\w)
\end{verbatim}\end{minipage} \\
\midrule
    Age & \begin{minipage}{13cm}\begin{verbatim}\b(?:I am|I'm|Im|I'?m|aged|age)\s*
(?P<age1>\d{2})(?!%)\s*(?:years? old|y\.?o\.?|years? of age)?\b |
(?<!\w)(?P<age4>\d{2})(?P<gender>(M|F|NB|AMAB|AFAB|MtF|FtM|
Enby|GM|GNC|TG|T|Cis))(?!\w)| 
(?<!\w)(?P<gender2>(M|F|NB|AMAB|AFAB|MtF|FtM|Enby|GM|GNC
|TG|T|Cis))(?P<age5>\d{2})(?!\w)
\end{verbatim}\end{minipage} \\
\midrule
Hobby & \begin{minipage}{13cm}\begin{verbatim}\b(?:I like|I enjoy|I love|I hate|I often|I usually|I prefer|I dislike|
I can't stand|I adore|I'm passionate about|I'm fond of|I'm interested
in|I'm into|I tend to|I have a habit of)\s+
(?P<preference>.+?)(?=[,.!?]|$)\end{verbatim}\end{minipage} \\
\midrule
Possession & \begin{minipage}{13cm}\begin{verbatim}\b(?:I have|I've got|I own|I possess|I hold|I acquired|I received)\s+
(?P<experience>.+?)(?=[,.!?]|$)\end{verbatim}\end{minipage} \\
\midrule
Work & \begin{minipage}{13cm}\begin{verbatim}\b(?:I work|I'm working|I'm employed|I have a job|I do work|I freelance
|I have been working|I started working|I used to work|
I studied | I am studying | I'm studying | Im studying | I have studied)\s+
(?P<job_details>.+?)(?=[,.!?]|$)
\end{verbatim}\end{minipage} \\
\midrule
Attitude & \begin{minipage}{13cm}\begin{verbatim}\b(?:I believe|I think|I feel|I support|I oppose|I stand for|I stand 
against|I'm against|I'm for|I'm pro-|I'm anti-|I value|I don't believe 
in|I consider|I advocate for|I reject|I agree with|I disagree with|
I am passionate about|I'm critical of|I align with|I side with|
I view|I respect|I distrust|I question|I doubt|I condemn|I appreciate|
I prioritize|I favor|I disapprove of|I endorse|I subscribe to|
I'm skeptical of)\s+
(?P<opinion>.+?)(?=[,.!?]|$)\end{verbatim}\end{minipage} \\
\midrule
Relationship & \begin{minipage}{13cm}\begin{verbatim}
\b(?:I am|I'm|Im|My|Our|We are|I have)(a|an)?\s*(?P<relationship>
mother|father|mom|dad|brother|sister|son|daughter|uncle|aunt|
grandfather|grandmother|grandpa|grandma|cousin|nephew|niece|
husband|wife|boyfriend|girlfriend|partner|spouse|
best friend|friend|roommate|fiancé|fiancée|in-law|step(?:mother|
father|brother|sister|son|daughter))\b\end{verbatim}\end{minipage} \\

    \bottomrule
    \end{tabular}
    \caption{Regular expressions used to match sentences of each type}
    \label{tab:regex}
\end{table*}

When doing k-means clustering, we use only posts that have a sentence that matches one of the following substrings, so that clustering has more to do with the personal statements:

\begin{itemize}
    \setlength\itemsep{0em}
    \item I am
    \item I'm
    \item Im
    \item I have
    \item I like
    \item I love
    \item I hate
    \item I enjoy
    \item I think
    \item I feel
    \item I believe
    \item I wish
    \item I need
    \item I want
    \item I fear
    \item I worry
    \item I tend to
    \item I see myself as
    \item I value
    \item I strive to
    \item I consider myself
    \item I would describe myself as
    \item I would define myself as
    \item I pride myself on
    \item I am good at
    \item I struggle with
    \item I find it easy to
    \item I have a hard time
    \item I excel at
    \item I know that I
    \item Ive learned that I
    \item I've learned that I
    \item I have learned that I
    \item I realize that
    \label{fig:sd-phrases}
\end{itemize}

\section{Self-disclosure Exploration}\label{sec:self_disclosure_exploration}

\begin{figure}[h]
    \centering
    \includegraphics[width=0.4\textwidth]{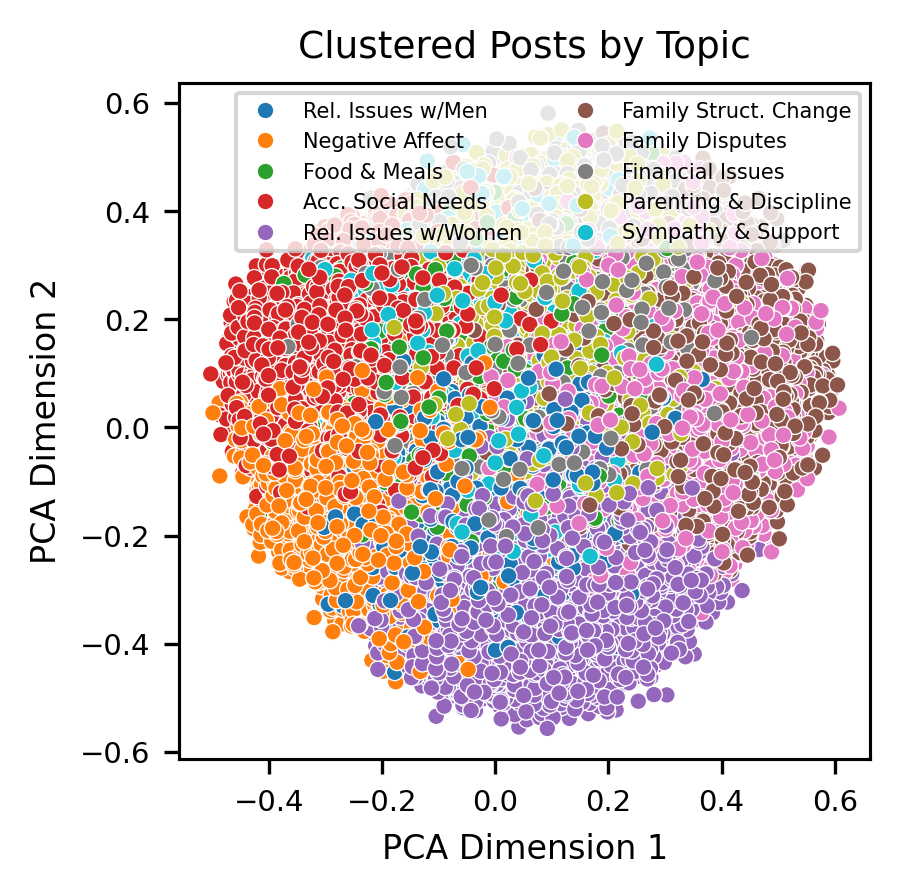}
    \caption{Results of kMeans clustering algorithm}
    \label{fig:kmeans_cluster}
\end{figure}

\begin{figure}[h]
    \centering
    \includegraphics[width=0.4\textwidth]{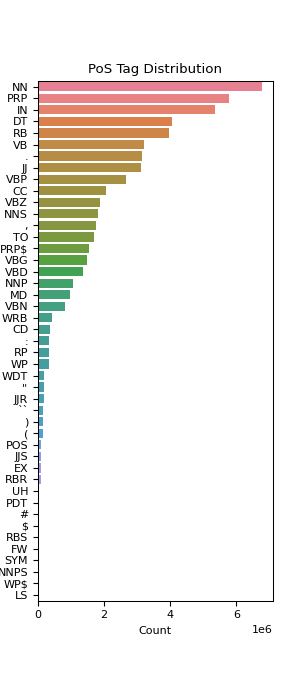}
    \caption{Distribution of Parts of Speech in posts containing self-disclosure phrases}
    \label{fig:pos}
\end{figure}

\begin{figure}[h]
    \centering
    \includegraphics[width=0.4\textwidth]{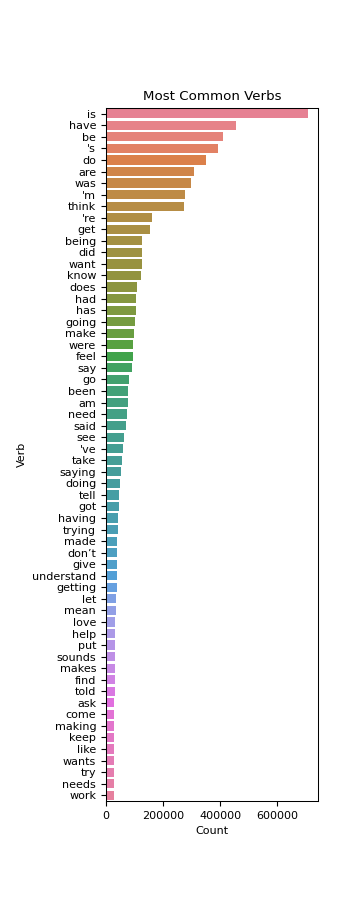}
    \caption{Distribution of verbs in posts containing self-disclosure phrases}
    \label{fig:verbs}
\end{figure}

\begin{figure}[h]
    \centering
    \includegraphics[width=0.4\textwidth]{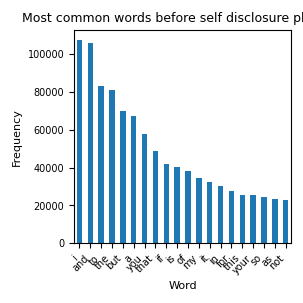}
    \caption{Most common unigrams before a self-disclosure phrase}
    \label{fig:before-unigram}
\end{figure}

\begin{figure}[h]
    \centering
    \includegraphics[width=0.4\textwidth]{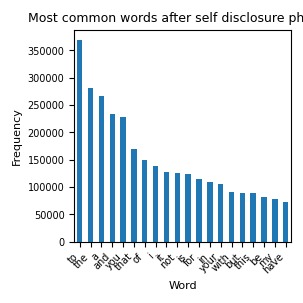}
    \caption{Most common unigrams after a self-disclosure phrase}
    \label{fig:after-unigram}
\end{figure}

\begin{figure}[h]
    \centering
    \includegraphics[width=0.4\textwidth]{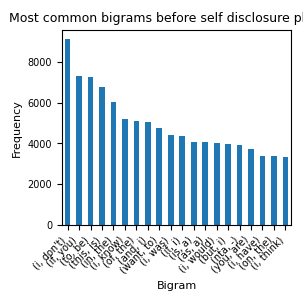}
    \caption{Most common bigrams before a self-disclosure phrase}
    \label{fig:before-bigram}
\end{figure}

\begin{figure}[h]
    \centering
    \includegraphics[width=0.4\textwidth]{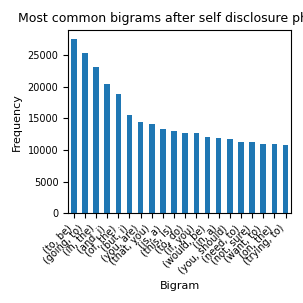}
    \caption{Most common bigrams after a self-disclosure phrase}
    \label{fig:after-bigram}
\end{figure}

\begin{figure}[h]
    \centering
    \includegraphics[width=0.4\textwidth]{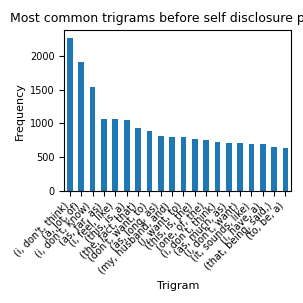}
    \caption{Most common trigrams before a self-disclosure phrase}
    \label{fig:before-trigram}
\end{figure}

\begin{figure}[h]
    \centering
    \includegraphics[width=0.4\textwidth]{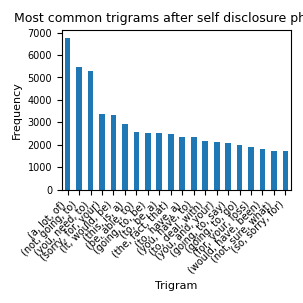}
    \caption{Most common trigrams after a self-disclosure phrase}
    \label{fig:after-trigram}
\end{figure}

In this appendix, we discuss additional explorations of the self-disclosure statements. We also analyzed the most common unigrams, bigrams, and trigrams around self-disclosure phrases, grouped authors by post count, tagged parts of speech to identify frequent patterns and verbs. When examining unigrams, bigrams, and trigrams, we looked at every word in a sentence before and after a self-disclosure phrase and identified the most common unigram, bigram, and trigrams for each. We found that the most common unigram before a self-disclosure is ``i'' (Fig.~\ref{fig:before-unigram}), and is ``to'' after (Fig.~\ref{fig:after-unigram}). The most common bigram before a self-disclosure is ``i don't'' (Fig.~\ref{fig:before-bigram}), and is ``to be'' after (Fig.~\ref{fig:after-bigram}). The most common trigram before a self-disclosure phrase is ``i don't think'' (Fig.~\ref{fig:before-trigram}), and is ``a lot of'' after (Fig.~\ref{fig:after-trigram}). The three most common Parts of Speech are NN, PRP, and IN (Fig.~\ref{fig:pos}). The three most common verbs are ``is'', ``have'', and ``be'' (Fig.~\ref{fig:verbs}). 
The most common self-disclosure phrases were ``I'm,'' ``I have'', and ``I am''. The most common verbs are ``was'', ``is'', and ``have''. The most common parts of speech are nouns, personal pronouns, and Preposition or subordinating conjunction. 
Additionally, we used PCA to reduce the dimension of our SBERT embeddings, from 384 (SBERT embedding dimension) to two, to plot the kmeans clusters ~\ref{fig:kmeans_cluster}. The PCA variance ratio for the first dimension is 0.043 and for the second dimension is 0.034.

\begin{figure}[h]
    \centering
    \includegraphics[width=0.4\textwidth]{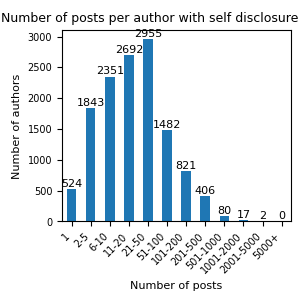}
    \caption{Number of posts made by each author}
    \label{fig:posts_per_author}
\end{figure}

\begin{table}[]
    \small
    \centering
    \begin{tabular}{lccc}
        \toprule
        \textbf{Annotator Model} & \textbf{5+\%} & \textbf{Accuracy} & \textbf{F1} \\
        \midrule
        \multicolumn{4}{c}{Theory Based High-level Categorization} \\ 
        \midrule
        Attitudes  & 68 & 85.9 & 83.2 \\ 
        Demographics  & 69 & 85.8 & 83.0 \\ 
        Experiences & 48 & 85.8 & 83.1 \\ 
        Relationships & 56 & 85.8 & 83.0 \\ 
        \midrule
        \multicolumn{4}{c}{Automatic High-level Categorization} \\ 
        \midrule
        C1: Financial Issues & 30 & 85.9 & 83.1 \\ 
        C2: Acc. Social Needs & 42 & 85.8 & 83.1 \\ 
        C3: Parenting \& Discipline & 33 & 85.9 & 83.1 \\ 
        C4: Family Struct. Change & 33 & 85.9 & 83.2 \\ 
        C5: Rel. Issues w/Men & 37 & 85.8 & 83.1 \\ 
        C6: Rel. Issues w/Women & 42 & 85.8 & 83.0 \\ 
        C7: Sympathy \& Support & 24 & 85.9 & 83.3 \\ 
        C8: Family Disputes & 32 & 85.8 & 83.1 \\ 
        C9: Negative Affect & 39 & 85.8 & 83.1 \\ 
        C10: Food \& Meals & 23 & 85.7 & 83.0 \\ 
        \bottomrule
    \end{tabular}
    \caption{Results for both theory based and automatic categorization of posts in the verdict split. 5+\% denotes the percentage of annotators that had at least 5 posts for each category.}
    \label{tab:category_results}
\end{table}

\begin{table}[]
    \small
    \centering
    \begin{tabular}{lccc}
        \toprule
        \textbf{Annotator Model} & \textbf{5+\%} & \textbf{Accuracy} & \textbf{F1} \\
        \midrule
        \multicolumn{4}{c}{Theory Based High-level Categorization} \\ 
        \midrule
        Attitudes  & 68 & 85.5 & 82.5 \\ 
        Demographics  & 69 & 85.4 & 82.5 \\ 
        Experiences & 48 & 85.4 & 82.6 \\ 
        Relationships & 56 & 85.5 & 82.6 \\ 
        \midrule
        \multicolumn{4}{c}{Automatic High-level Categorization} \\ 
        \midrule
        C1: Financial Issues & 30 & 85.6 & 82.7 \\ 
        C2: Acc. Social Needs & 42 & 85.3 & 82.6 \\ 
        C3: Parenting \& Discipline & 33 & 85.4 & 82.6 \\ 
        C4: Family Struct. Change & 33 & 85.5 & 82.5 \\ 
        C5: Rel. Issues w/Men & 37 & 85.5 & 82.5 \\ 
        C6: Rel. Issues w/Women & 42 & 85.5 & 82.6 \\ 
        C7: Sympathy \& Support & 24 & 85.4 & 82.5 \\ 
        C8: Family Disputes & 32 & 85.4 & 82.5 \\ 
        C9: Negative Affect & 39 & 85.4 & 82.5 \\ 
        C10: Food \& Meals & 23 & 85.4 & 82.6 \\ 
        \bottomrule
    \end{tabular}
    \caption{Results for the author split data}
    \label{tab:author_results}
\end{table}

\section{Other Splits}\label{sec:appendix_other_splits}

To evaluate how well models generalize across different conditions, we adopt three data splitting strategies used in prior work \cite{plepi-etal-2022-unifying}, each targeting a specific type of generalization. We computed results for the individual categories, both automatic and theory-based. The situation split, which proved to be the most difficult in previous work, is in the main body of the paper. The verdict split is shown in Table~\ref{tab:category_results} and the results for the author split are shown in Table~\ref{tab:author_results}.

\mysubsection{Verdict split}. This split randomly assigns verdicts to the training, validation, and test sets. While straightforward, it introduces two potential sources of overlap: the same situations and the same annotators may appear in multiple splits. Our dataset forms a fully connected bipartite graph between posts and annotators, where each edge represents a comment. Because many annotators comment on many posts, it's not possible to remove both sources of overlap at the same time without disconnecting the graph or discarding large portions of data. This split is equivalent to the ``no disjoint'' setting described in \cite{plepi-etal-2022-unifying}.

\mysubsection{Situation split}. To test how well models generalize to new situations, we ensure that the posts (situations to assign a verdict to) in the training, validation, and test sets are completely disjoint. Annotators may still appear in more than one split.

\mysubsection{Author split}. This setup evaluates how well models generalize to new annotators. Here, each annotator appears in only one of the train, validation, or test sets, while posts may be shared across them.
\newline
\noindent
Each of these splits controls for a different variable. Together, they allow us to assess how self-disclosure information contributes to predicting judgments across new contexts and individuals.

\section{Discussion of Automatic Cluster Labels}

To identify each automatic cluster, two of the authors independently inspected the 5 comments closest to the centroids of each cluster along with 5 random comments from each cluster to label each cluster. The group then deliberated and made a final decision on the titles together. Below is a more in depth exploration of the reasoning behind each cluster with example comments. For each cluster we also provide an example comment from the centroid of the cluster and a random comment from the cluster.

\subsection{C1: Financial Issues}
The \textit{Financial Issues} cluster contains comments which discuss topics like money, budgets, loans, or directly mention the cost of something, for example \$20. 

\textbf{Centeroid Comment:}
    ``NTA. I totally get both sides. I feel uncomfortable when people spend too much money on me too. It's a pride thing I guess. You work your tail off and still aren't in a solid financial position.
    ...
    I know where I'm at, \$1,200 is around what 2.5 weeks of rent + utilities for a family of 5 would cost anyway, you're just paying it to her instead of a landlord.''

\textbf{Random Comment:}
``NTA - please DO NOT sell that car! I think you'll regret it if you do. You love your sister and want to help, but it's her choice to study abroad and she needs to find a way to make it happen. Why should you have 
to sacrifice something so special just because you share dna and she wants money.''

\subsection{C2: Accommodating Social Needs }
The \textit{Accommodating Social Needs} cluster contains comments which discuss topics centered around Accommodating others, social obligation, and social norms.

\textbf{Centeroid Comment:}
``YTA. I'm like this. Partner of many years had never been annoyed by it (he even helps me keep track of where the nearest bathrooms are and asks if I need to go before we leave)''

\textbf{Random Comment:}
``I'm glad I could help. I hope you find healing.''

\subsection{C3: Parenting \& Discipline}
The \textit{Parenting \& Discipline} cluster contains comments which discuss topics like general parenting guidelines/styles, disciplining children, and advice for parents.

\textbf{Centeroid Comment:}
``I think you may be downplaying how much a parent can ignore or not see their own child’s bad behavior. I have a feeling that’s what is going on here.''

\textbf{Random Comment:}
``... I have literally NEVER witnessed any new parents doing something so selfish and self-absorbed. Like I said, 
and like you said, it's THEIR baby and they have every right to do whatever they want. But it WILL damage their relationships with the people they deem `close' to them.''

\subsection{C4: Family Structure Change}
The \textit{Family Structure Change} cluster contains comments which discuss topics mainly centered around changes in the family, either adding or subtracting in some way. Examples of this include moving away, divorce, death, or birth in the family which changes how the family operates.

\textbf{Centeroid Comment:}
``Love how OP tries to frame it as "looking out for his son" when in reality, he selfishly wants his son to be a doormat \& do what he wants versus being happy.  I'm betting the daughter got the same treatment \& that's why she went NC to avoid all his stupid drama.
...
They've grown up \& learned to be 
adults, about damn time you do the same before you torpedo any chances of ever having them in your lives again. 
...''

\textbf{Random Comment:}
``Husband has been clear about not wanting another baby but OP has insisted on carrying to term. I think he's been clear about not wanting to parent.''

\subsection{C5: Relationship Issues with Men}
The \textit{Relationship Issues with Men} cluster contains comments which discuss situations in which people have problems with men.

\textbf{Centeroid Comment:}
``I think you and him need to have a serious conversation about this. That behavior is unacceptable.''

\textbf{Random Comment:}
``...
In short, you should be able to watch your CNN at his house and he should be able to watch his Fox News. This has nothing to do with his cancer diagnosis, nor does it have to do specifically with your relationship to him.
...''

\subsection{C6: Relationship Issues with Women}
The \textit{Relationship Issues with Women} cluster contains comments which discuss situations in which people have problems with women.

\textbf{Centeroid Comment:}
``NTA that sounds a little mean to be honest. I think it's understandable to be upset by it, but hopefully she will take it seriously and stop doing that.''

\textbf{Random Comment:}
``...
I would NEVER talk to my boyfriend that way for trying to help me. You were right to leave. You were being an excellent boyfriend, but she was being a jerk.''

\subsection{C7: Sympathy \& Support }
The \textit{Sympathy \& Support} cluster contains comments which offer support and sympathy for the original poster. Almost all of the comments in this cluster contain ``Not The Asshole'' (NTA).

\textbf{Centeroid Comment:}
``Definitely NTA. I'm sorry you have to put up with that kind of behavior.''

\textbf{Random Comment:}
``NTA. I'm just amazed that your sister is a 30 year old woman acting like that.''

\subsection{C8: Family Disputes }
The \textit{Family Disputes} cluster contains comments which discuss issues between family members. This usually occurs as arguments or somebody feeling slighted by another family member and asking Reddit for advice. This cluster is different from C4, \textit{Family Structure Change} because it does not focus on a change in the structure of the family and rather on the issues and arguments between family members.

\textbf{Centeroid Comment:}
``ESH, apart from the sister. It's understandable that you have these feelings, however, you're portraying your sister as if she has the capacity to understand what she's doing and does it to spite you, which makes 
you TA. Your parents sounds like they have neglected you, which is not ok, it was probably unintentional, but even so, it's obviously had a deep impact on you. Your Dad's comment was way out of line.''

\textbf{Random Comment:}
``And your grandparents helped perpetuate this hurtful fraud. Fuck you really didn't have anyone in your corner, it must have felt like you against the world. I am so sorry your family are such selfish hurtful fucks. NTA.''

\subsection{C9: Negative Affect}
The \textit{Negative Affect} cluster contains comments in which people are disagreeing with something or talking about civility.

\textbf{Centeroid Comment:}
``I disagree with that. I like to judge the totality of a situation and not a mere instance. ...''

\textbf{Random Comment:}
``... I was objecting to the NAH I was replying to.''

\subsection{C10: Food \& Meals }
The \textit{Food \& Meals} cluster contains comments which discuss topics related to food and eating.

\textbf{Centeroid Comment:}
``...
You were my nightmare as a child. I was a terribly picky eater as a kid and going to sleepovers at friends' houses was always a gamble because I was scared to try their food. Some people don't like your food, that doesn't make them an asshole, it just makes you not that great a cook.
...''

\textbf{Random Comment:}
``YTA, I’m a chef and even I hate cooking dinner at home. I have two children so I cook when they are with me. It’s a lot harder to even cook for two people than it is for one. Double time preparing what ever ingredients are needed and probably more washing up at the end also. ...''

\end{document}